\documentclass[10r pt]{article}
\usepackage[utf8]{inputenc}
\usepackage{natbib} 
\usepackage{fullpage}
\usepackage{amsmath}
\usepackage{amssymb}
\usepackage{cleveref} 
\usepackage{graphicx}
\usepackage{caption} 
\usepackage{subcaption}
\usepackage{cleveref}
\usepackage{url}
\newcommand{\op}{\operatorname}
\makeatletter
\newcommand\footnoteref[1]{\protected@xdef\@thefnmark{\ref{#1}}\@footnotemark}
\makeatother
\usepackage[Algorithm,ruled]{algorithm}
\usepackage[noend]{algpseudocode} 

\newcommand{\context}{w}
\title{A Note on Task-Aware Loss via Reweighing Prediction Loss by Decision-Regret}
 \author{ Connor Lawless\footnote{Authors listed in alphabetical order.}\\
 Cornell University \and 
Angela Zhou${^*}$ \\ USC Marshall School of Business }
\date

\begin{document}

\maketitle

\begin{abstract}
In this short technical note we propose a baseline for decision-aware learning for contextual linear optimization, which solves stochastic linear optimization when cost coefficients can be predicted based on context information. We propose a decision-aware version of predict-then-optimize. We reweigh the prediction error by the decision regret incurred by an (unweighted) pilot estimator of costs to obtain a \textit{decision-aware predictor}, then optimize with cost predictions from the decision-aware predictor. This method can be motivated  as a finite-difference, iterate-independent approximation of the gradients of previously proposed end-to-end learning algorithms; it is also consistent with previously suggested intuition for end-to-end learning. This baseline is computationally easy to implement with readily available reweighted prediction oracles and linear optimization, and can be implemented with convex optimization so long as the prediction error minimization is convex. Empirically, we demonstrate that this approach can lead to improvements over a "predict-then-optimize" framework for settings with misspecified models, and is competitive with other end-to-end approaches. Therefore, due to its simplicity and ease of use, we suggest it as a simple baseline for end-to-end and decision-aware learning. 
\end{abstract}

\section{Introduction}

Data-driven optimization solves optimization problems with stochastic cost inputs (ex. predictable, but random edge costs on a network). Machine learning and statistics predict stochastic outcomes from covariate data. Contextual stochastic optimization solves stochastic optimization problems where the stochastic cost inputs can be predicted from auxiliary data. In the case of linear optimization, linearity of expectation implies that consistent predictors are sufficient to achieve consistent decision regret. However, there remains great interest in other approaches, because data-driven-optimization seeks good \textit{decision regret} of the final decisions made in view of predictions, while standard prediction approaches target the prediction error rather than the decision-regret. Approaches and proposed refinements fall into a few general categories: ``predict then optimize", optimizing based on predictions alone, vs. decision-aware approaches. Decision-aware approaches include ``end-to-end" learning approaches which attempt to optimize the task loss directly \citep{donti2017task,wilder2019melding}, differentiating through the dependence of the decision regret on parameters (e.g. argmin differentiation); as well as other modifications to predict-then-optimize such as deriving surrogate losses to the decision regret \citep{el2019generalization}. 

In this short note we propose re-weighting \textit{prediction mean-squared error (MSE)} by the decision regret (obtained by a pilot estimate) as a natural baseline for contextual linear stochastic optimization. At its face, it accords with a key intuition offered in the literature to justify end-to-end approaches for linear optimization: \textit{incorporating decision-regret information improves prediction accuracy in covariate regions that have greater effect on decisions.} It can also be justified as a zeroth-order finite difference, and iterate-independent, approximation to the gradients of the decision-aware end-to-end loss. Since re-weighing prediction error by the decision regret does not require any algorithmic modifications -- it can be implemented via prediction and linear optimization oracles -- it is a computationally appealing baseline. 

Our contributions are as follows. We propose a method which learns a decision-aware predictor by reweighing the squared loss by the decision-regret of a pilot (i.e. unweighed) predictor. We show in preliminary empirical experiments that its performance is comparable in a small motivating example and in experiments with the SPO+ decision-aware loss. On the other hand, we also observe that this approach suffers typical issues with reweighed empirical risk minimization; for nonparametrically rich model such as random forests, reweighing prediction loss tends not to change the predictions. Reweighing prediction loss by decision-regret combines two desiderata: decision-awareness, and computational tractability. We argue it is a reasonable baseline.


\section{Problem Setup and Related Work} 
\subsection{Problem Setup} 

We consider a linear program, $\max \{ c^\top x \colon x \in \mathcal X\}$. The decision variable $x \in \mathcal X$ is restricted to the $d-$ dimensional polytope $\mathcal X$. In the stochastic optimization setting with contextual costs, the objective function coefficients $c$ are in fact random variables, measurable with respect to covariates $z \in \mathbb{R}^p$ with conditional expectation $\mathbb E[c \mid z]$. We denote a generic cost predictor as $\hat c(z)$ and omit arguments of the function at times for brevity. We will generally consider predictors parametrized by some parameter $\theta \in \mathbb{R}^p$ and therefore denote the predictions parametrized by $\theta$ as $\hat{c}_\theta(z)$. 

At a new test instance, we observe a realization of $z$, then make a decision $x\in \mathcal{X}.$ We consider solutions from predict-then-optimize with a generic predictor $c$, denoted as the argmax mapping $x^*(c)$. ``Predict-then-optimize" generally refers to plugging in (decision-unaware) predictors, then optimizing with those predictions. We incur final costs of $c^\top x,$ i.e. depending on the realized demand. The decision regret incurred by a predictor $\hat c$ is the regret relative to the best solution under the true cost realizations, $x^*(c)$: 
\begin{equation} c^\top (x^*(\hat c) - x^*(c) )
\end{equation}

Our available data therefore comprises of $\{(z_i,c_i)\}_{i=1, \dots n}$ tuples of contexts and observed cost realizations. 
The (decision-unaware) least-squares prediction error, which is sufficient for regression, is 
 $\ell_{\op{MSE}}(\hat c_\theta, c) = (\hat c_\theta- c)^2  $ and the prediction loss gradient is $\nabla_\theta \ell_{\op{MSE}}(\hat c, c) = 2(\hat c_\theta -c) \frac{\partial \hat c_\theta}{\partial \theta} $. 

\subsection{Related work} 
We briefly discuss decision-aware approaches in greater detail, and refer to \cite{kotary2021end} for survey material on these methods.

\paragraph{End-to-end learning}
We first describe approaches based on end-to-end learning which differentiate through the optimality conditions of the optimization program, as pursued in \citep{donti2017task,wilder2019melding} and other works; as we will later compare gradients of our loss function with the gradients in end-to-end learning.

Consider a generic optimization problem $\max\{ f(x, \theta) \colon {x \in \mathcal{X}}
\}.
$ For example, for linear programs, $f(x,c)=c^\top x.$
End-to-end learning or task loss approaches optimize the objective value attained by a parametrized predictor $c_\theta$ with respect to the parameter $\theta$ \citep{kotary2021end}. The task loss is precisely the objective function of the optimization at this solution $x^*(c)$ induced by a generic predictor $c$. 

End-to-end approaches directly optimize the expected task loss $\mathbb E[f(x^*(\hat{c}_\theta),c)]$ with respect to prediction parameter $\theta$. They minimize the expected loss by computing the following gradient: 
\begin{equation}
\frac{d f(x^*(\hat{c}_\theta), c)}{d \theta}=\frac{d f(x^*(\hat{c}_\theta), c)}{d x^*(\hat{c}_\theta)} \frac{d x^*(\hat{c}_\theta)}{d \hat{c}_\theta} \frac{d \hat{c}_\theta}{d \theta} \label{eqn-taskloss}
\end{equation}

For linear objectives, the first term is the cost vector $\frac{d f(x^*(\hat{c}), c)}{d x^*(\hat{c})} = c$, the middle term is the implicit function/argmin differentiation term $\frac{d x^*(\hat{c})}{d \hat{c}}$, and the last term $\frac{d \hat{c}}{d \theta}$ is the  gradient of predictions with respect to parameters $\theta$. The middle term is computed by ``argmin differentiation", i.e. differentiating through the optimality conditions. It is specific to these approaches for decision-aware end-to-end learning. 

\citep{donti2017task} leverages the quadratic-program-specialized argmin-differentiation of \citep{amos2017optnet}, which differentiates through the Karush-Kuhn Tucker conditions via the implicit function theorem, to directly optimize the task loss with respect to the prediction parameters, such as those of a deep net. \citep{donti2017task} focused on quadratic programs. \citep{wilder2019melding} considered the combinatorial optimization (mixed-integer linear program) case via linear program relaxations; \citep{ferber2020mipaal} considered cutting-plane algorithms. Both these papers add quadratic smoothing (squared norm of the decision vector) in order to leverage the same argmin differentiation.

\paragraph{Smart predict-then optimize}
\citep{elmachtoub2017smart} define the ``smart predict, then optimize" loss as the decision regret. They derive a surrogate loss of the nonconvex decision regret via inequalities obtained from properties of the optimal LP solution and additional approximation steps, which is $\ell_{\mathrm{SPO}+}(\hat{c}, c):=\max _{w \in S}\left\{c^T w-2 \hat{c}^T w\right\}+2 \hat{c}^T w^*(c)-z^*(c)$. Their approach results in a loss whose subgradient (with respect to $\hat c$) is $x^{*}(c)-x^{*}(2 \hat{c}-c)$. \citep{mandi2020smart} investigate extending SPO for combinatorial optimization problems, and find comparable performance in using simpler oracles such as LP relaxation of the comb. opt problem than full re-optimization. Further work studies learning-theoretic properties of the SPO+ loss \citep{el2019generalization,liu2021risk}. 

\section{Task-loss reweighted predictors}


These previously mentioned approaches have various decision-theoretic, statistical, and computational trade-offs. ``Predict-then-optimize" is consistent under well-specified regression predictions (and conversely, is vulnerable to model-misspecification), even though it is not specialized further to the decision regret setting. Decision-aware losses either require approximation arguments to derive a convex surrogate loss function (SPO+), else optimize a nonconvex loss (SPO); while end-to-end learning incurs computational difficulties of computing the implicit gradients. 

In this paper, we suggest task-loss reweighted predictors as a baseline. They further specialize least-squares prediction loss to the decision-aware setting by reweighting. But, they preserve the computational tractability of (weighted) empirical risk minimization. 

\paragraph{Decision-risk reweighted MSE} 
An initial proposal is to directly reweight the MSE by the decision regret achieved by the predictor (at the optimization parameter). 
\begin{equation}\label{eqn-dec-reweighted-mse}  \ell_{ oracle}( c_\theta,c) =  c^\top (x^*( c_\theta) - x^*(c)) ( c_\theta - c)^2 
\end{equation} 
This is consistent with various intuitive explanations for the success of task loss compared to prediction error given in the literature. However, optimizing this reweighted loss function with respect to $\theta$ introduces dependence of the decision loss weights on $\theta$.

We instead suggest a \textit{decision-aware predictor},  which optimizes the prediction risk reweighted by a feasible version of the decision-regret. We first describe how the feasible decision-regret weights are obtained. First, learn a pilot estimator $\hat c$ via unweighted mean squared error. 
We use the decision regret of $\hat c$ to reweigh the mean squared error, replacing $(x^*( c_\theta) - x^*(c))$ of \cref{eqn-dec-reweighted-mse} with $(x^*(\hat c) - x^*(c))$. Therefore, the decision regret used to reweigh the mean-squared error is \textit{independent} of the predictor parameters that we optimize currently. We obtain $\tilde c^*$ by reweighted error minimization. Our final output optimizes with respect to plug-in predictions from $\tilde c^*$. Because the weights are zero-inflated, since $x^*(\hat c) = x^*(c)$ if $\hat c$ is in the optimal basis for $c$,
we generally suggest to use mixture weights between unweighted squared loss and the decision-regret reweighted loss. These weights, denoted as $\nu$, can be tuned via cross-validation. 

To summarize, we obtain the following predictors:
\begin{align}
    &\hat c = c_{\hat\theta}, && \hat\theta \in \arg\min_\theta  \mathbb E[( c_\theta(z) - c)^2]&& \text{pilot estimate,} \nonumber\\
    &\tilde c^* = c_{\tilde{\theta}^*}, && \tilde{\theta}^* \in \arg\min_{ \theta} \mathbb E[(c^\top (x^*(\hat c) - x^*(c))) \cdot ( c_\theta(z) - c)^2] && \text{decision-aware predictor,}
    \label{eqn-feasible-dec-risk}
\end{align}
and solve the linear optimization problem with predictions from the \textit{decision-aware predictor} $\tilde c^*.$
This can be further generalized to an \textit{iteratively} reweighted approach in \Cref{alg-decision-reweighted-MSE}, i.e. replacing the pilot predictor with the previously obtained decision-aware predictor.
    
    \begin{algorithm}[t!]
    	\caption{Iteratively-reweighted decision-aware predict-then-optimize}\label{alg-decision-reweighted-MSE}
    	\begin{algorithmic}[1]
    	\State Input: $K$ rounds of reweighting, $\alpha_0=0$, $\nu\in(0,1)$ weight mixing hyperparameter.
    	\For{$k$ in $0...K$}
\State 
$\tilde \theta^*_k \in \underset{ \theta}{\arg\min}
        \;\; \mathbb E[  (\nu \alpha_k + (1-\nu) )  ( c_\theta- c)^2 ], \tilde c^*_k \gets c_{\tilde \theta^*_k}$
        \State 
        $\alpha_{k+1} \gets c^\top (x^*(\tilde c^*_k) - x^*(c))$.
        \EndFor
        \State Output $\tilde c^*_K$. Return the decision $x^*(\tilde c^*_K) \in \arg\max  \{ c^*_K x\colon x \in \mathcal X\}$. 
        \end{algorithmic} 
    \end{algorithm}

All the steps of this algorithm are implemented via readily available computational oracles: reweighted squared loss minimization, and optimization given a predictor.

\paragraph{Discussion and comparison}

Comparing the end-to-end approach (\cref{eqn-taskloss}) to ours (\cref{eqn-feasible-dec-risk}), note that for linear optimization, the \textit{gradients} of our feasible decision-risk reweighted loss are similar to the gradients of \cref{eqn-taskloss} but with the zero-order finite-difference approximation $x^*(\hat c) - x^*(c)$ based on a pilot estimator $\hat c$ that is independent of the iterates of the algorithm. Hence there are two sources of approximation: zero-order vs. first-order gradients, and a constant, iteration-independent approximation of the implicit differentiation term $\frac{d x^*(\hat c_\theta)}{d \hat c_\theta}.$ 

Despite these approximations relative to end-to-end learning, there are also some advantages of the feasible risk-reweighting approach. Our approach retains the well-behaved optimization properties of (reweighted) squared loss minimization. As long as the original squared loss minimization was convex, so too is our approach. Even if it is not, our approach is not computationally harder than the original ERM problem. 
Although the general formulation for task loss states to optimize \Cref{eqn-taskloss}, in practice (and in experiments that appear previously in the literature), it is suggested to initialize predictor weights by minimizing the standard prediction MSE, and then ``fine-tuned'' with the task loss\footnote{Personal communication.}. 

Note that in contrast to end-to-end loss in general, the population optimizer of the decision-risk reweighted loss (with positive mixture weight on the unweighted loss) is the Bayes-optimal prediction function. Therefore, in the asymptotic limit, in the well-specified case, our approach coincides with pure predict-then-optimize (which is indeed consistent). Decision-aware approaches such as end-to-end learning or SPO may be preferred in the case of model misspecification of the predictor function class. Our approach can be understood as another option in the face of potential model-misspecification. 

    
An analogy to our approach is the method of feasible generalized least squares (GLS) in econometrics \citep{takeshi1985advanced}. In the heteroskedastic regression setting, efficient generalized least squares arises from reweighting MSE by the conditional variance. Of course, the conditional variance is dependent on the the unknown prediction function (and its parameter). In GLS, if the error covariance is unknown, a consistent estimator of the weights may be obtained by using feasible GLS (FGLS), where a consistent pilot estimator is used to obtain a pilot estimate of the covariance matrix (and hence, the weights). FGLS is asymptotically efficient but finite sample properties are generally unknown.


\section{Illustrative Example}
\begin{figure}[t]
    \centering
    \begin{subfigure}[t]{0.45\linewidth}\includegraphics[width=\textwidth]{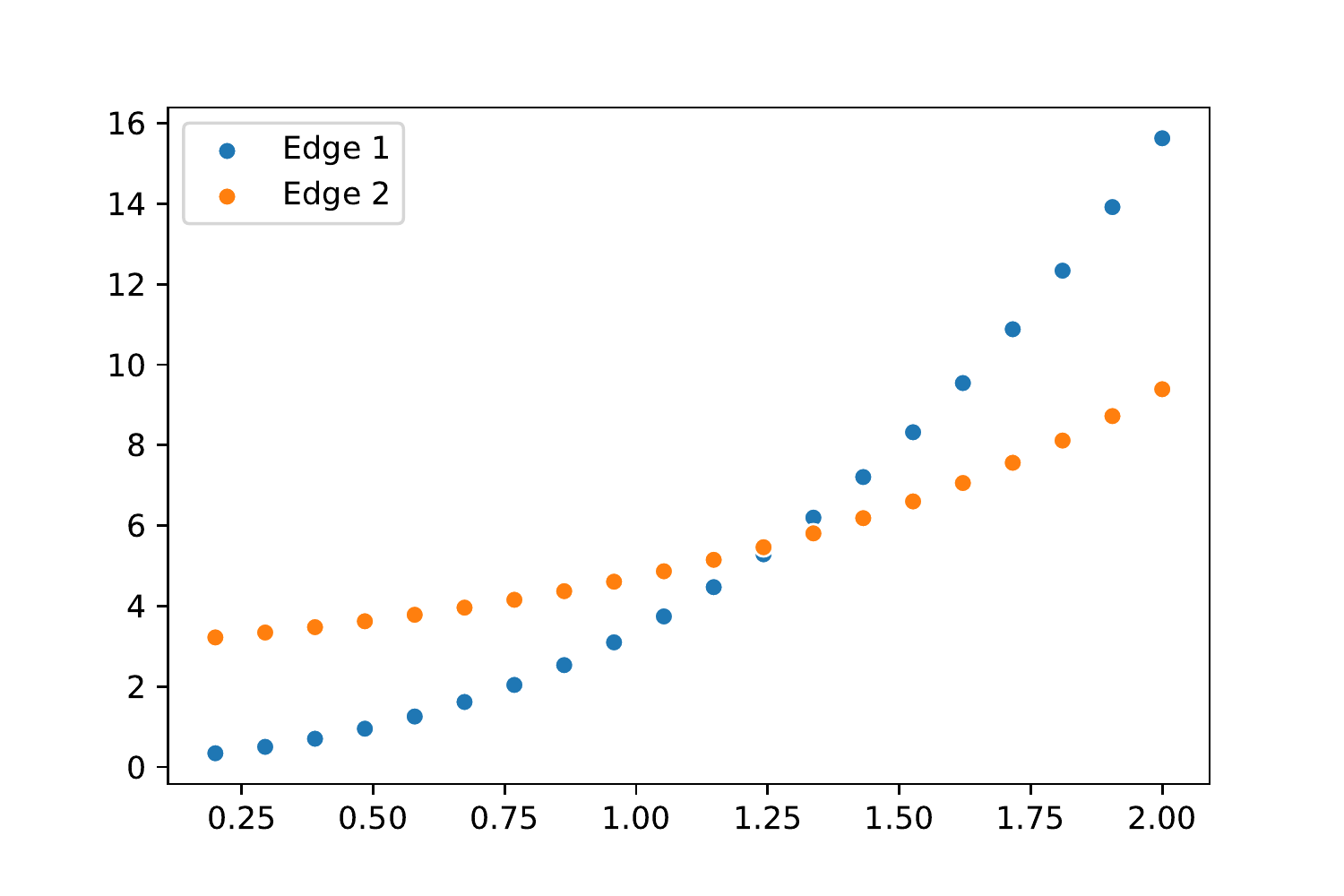}\end{subfigure}
    \begin{subfigure}[t]{0.45\linewidth}\includegraphics[width=\textwidth]{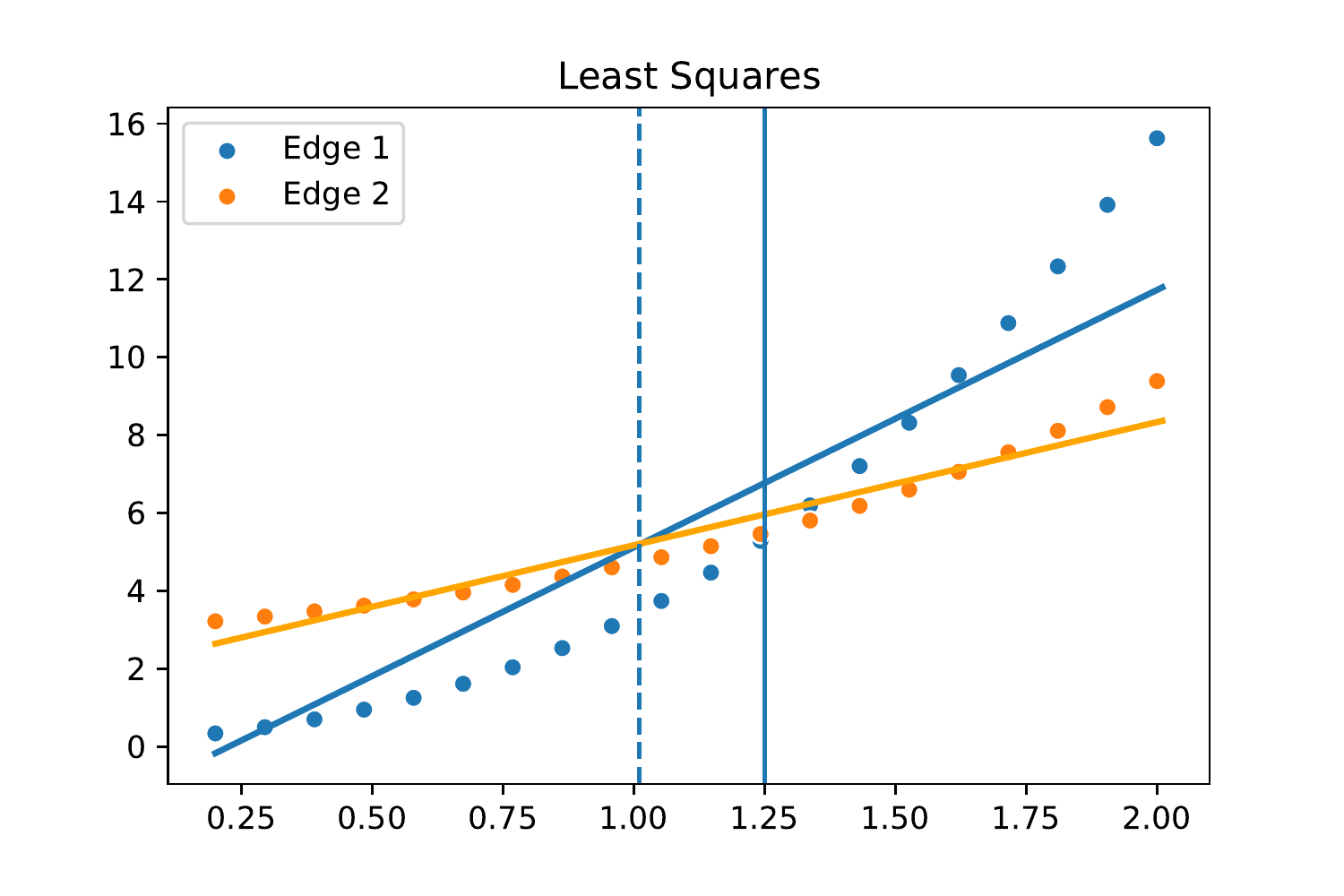}
    \end{subfigure}
    
     \begin{subfigure}[t]{0.45\linewidth}\includegraphics[width=\textwidth]{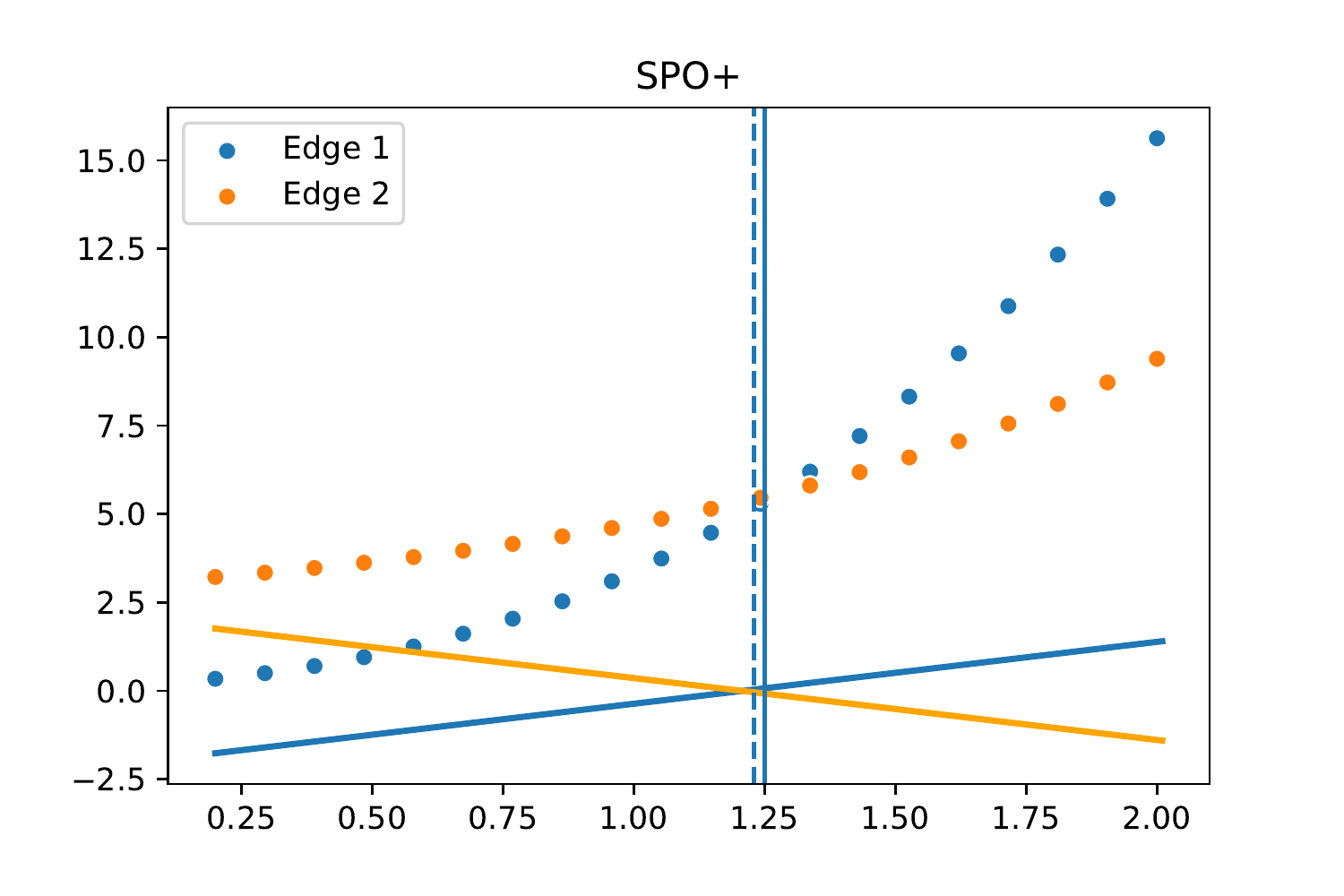}\end{subfigure}
     \begin{subfigure}[t]{0.45\linewidth}\includegraphics[width=\textwidth]{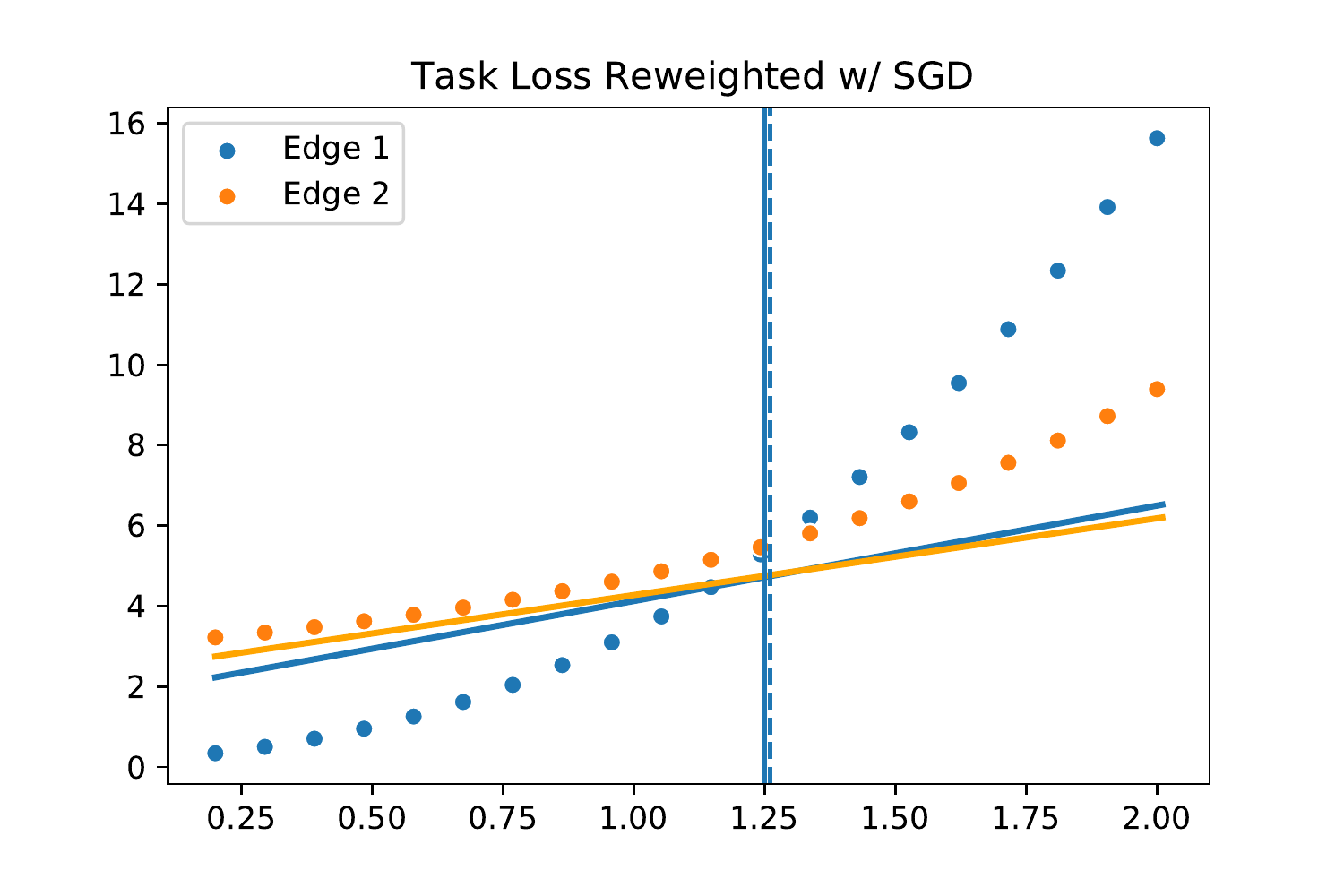}
    \end{subfigure}
    \caption{Simple 2 path shortest path problem. In this setting, we aim to fit a linear model to each edge, despite the true functions being decidedly non-linear. The top left plot shows the underlying edge costs for the training data points, and the other three graphs show the linear models trained under three approaches. }
    \label{fig:illustrative example}
\end{figure}

To help build some intuition behind our approach, we revisit an illustrative example given in \cite{elmachtoub2017smart} to highlight decision-aware learning vs predictive estimation. Consider a shortest path problem with two edges, and one feature that can be used to predict the edge costs. Figure \ref{fig:illustrative example} (top left) illustrates. Note that the underlying relationship between the feature and the edge costs are non-linear. When we try to fit a linear regression to each edge cost using standard least squares (top right plot) -- we get that the decision boundary (the dotted line) between selecting edge 1 vs edge 2 is very different from the true decision boundary (full line). However, SPO \citep{elmachtoub2017smart} (bottom left plot) is able to learn linear models that correctly predict the true decision boundary, but have edge costs that remain relatively divorced from the true dynamic (i.e. the slope for edge 2 is a different sign than the true edge 2 model). 

\begin{figure}[t]
    \centering
    \begin{subfigure}[t]{0.48\linewidth}\includegraphics[width=\textwidth]{illustrative_ex_figs/base_set-up.pdf}\end{subfigure}
    \begin{subfigure}[t]{0.48\linewidth}\includegraphics[width=\textwidth]{illustrative_ex_figs/mse.pdf} \end{subfigure}
    \begin{subfigure}[t]{0.48\linewidth}\includegraphics[width=\textwidth]{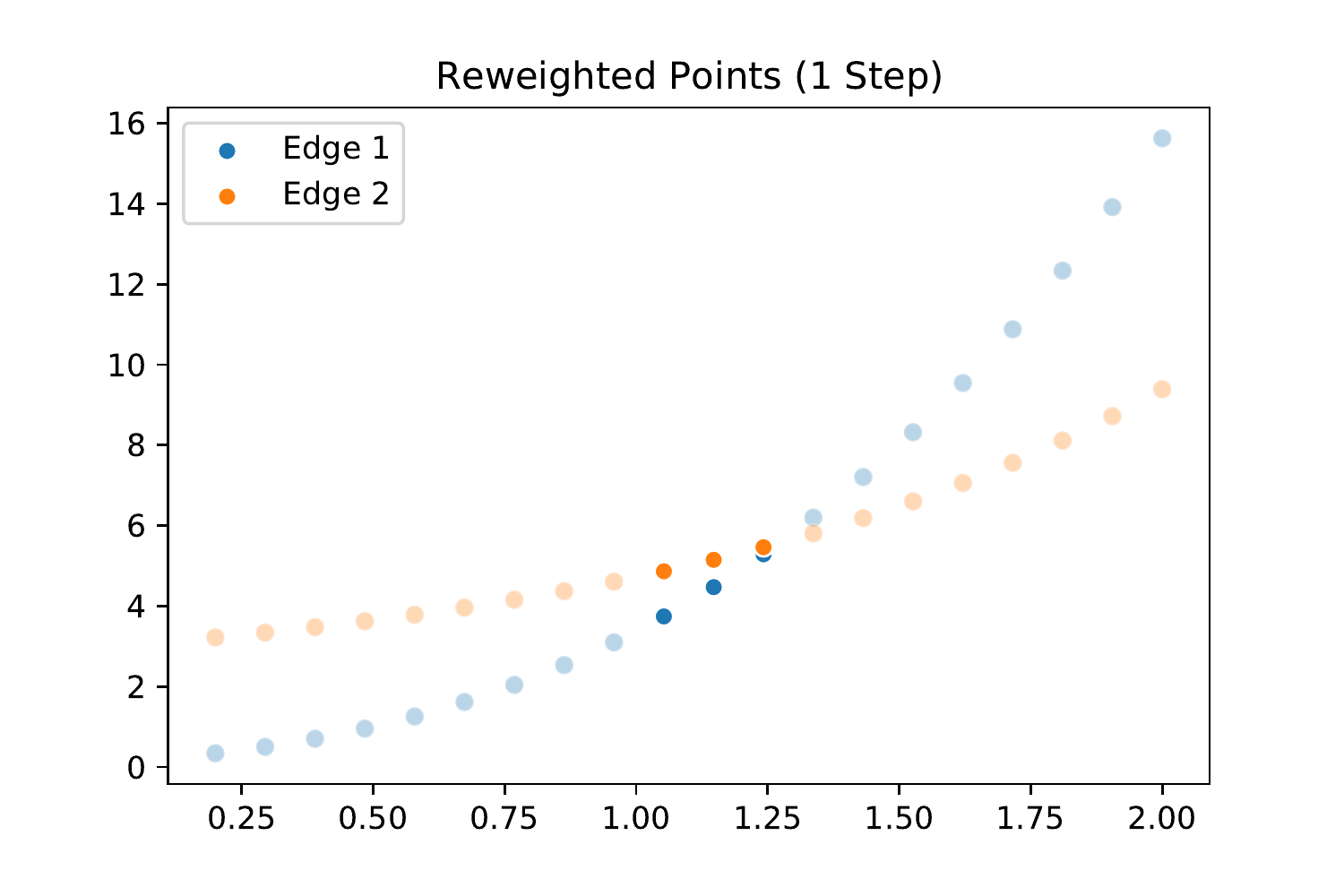}\end{subfigure}
    \begin{subfigure}[t]{0.48\linewidth}\includegraphics[width=\textwidth]{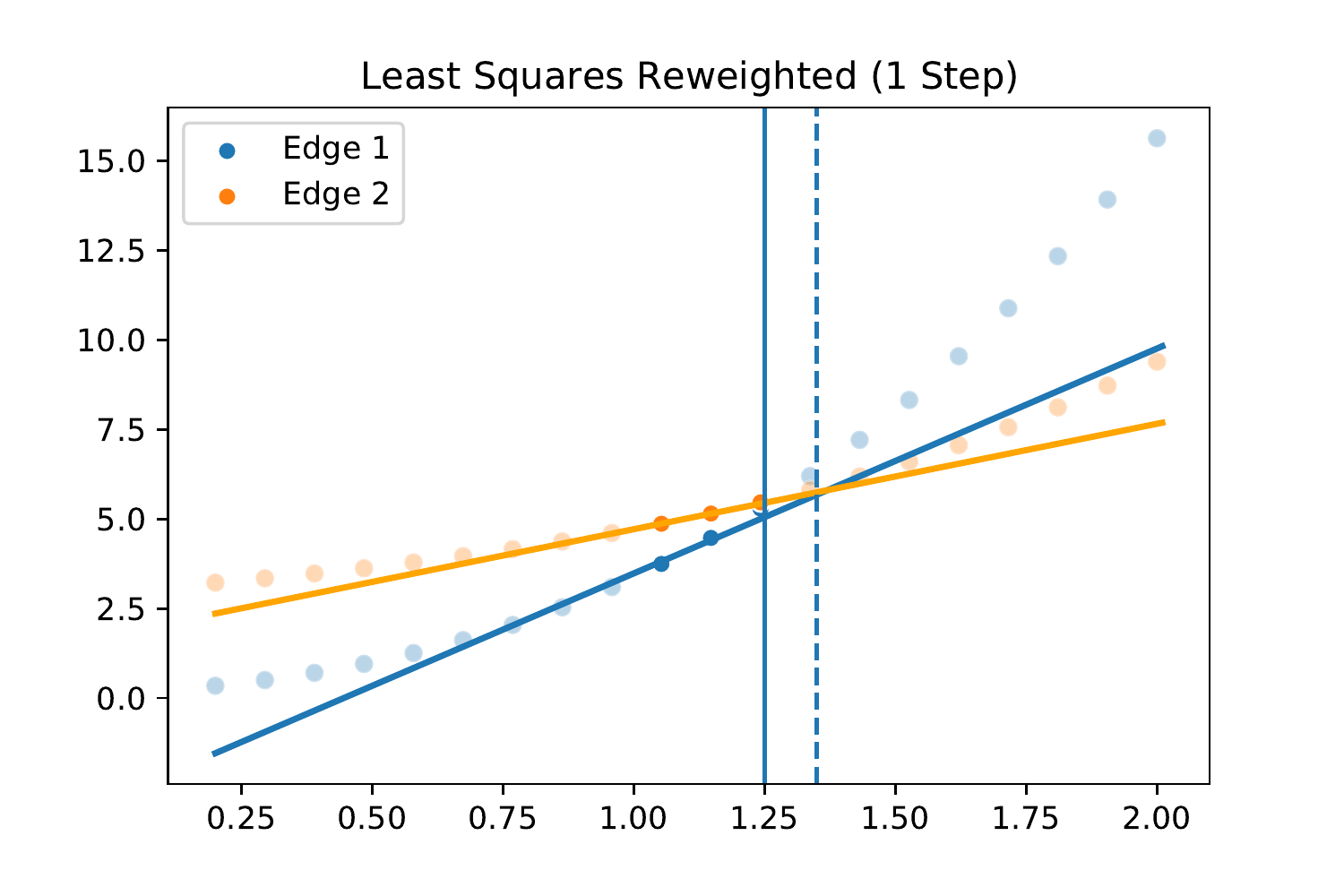}
    \end{subfigure}
     \begin{subfigure}[t]{0.48\linewidth}\includegraphics[width=\textwidth]{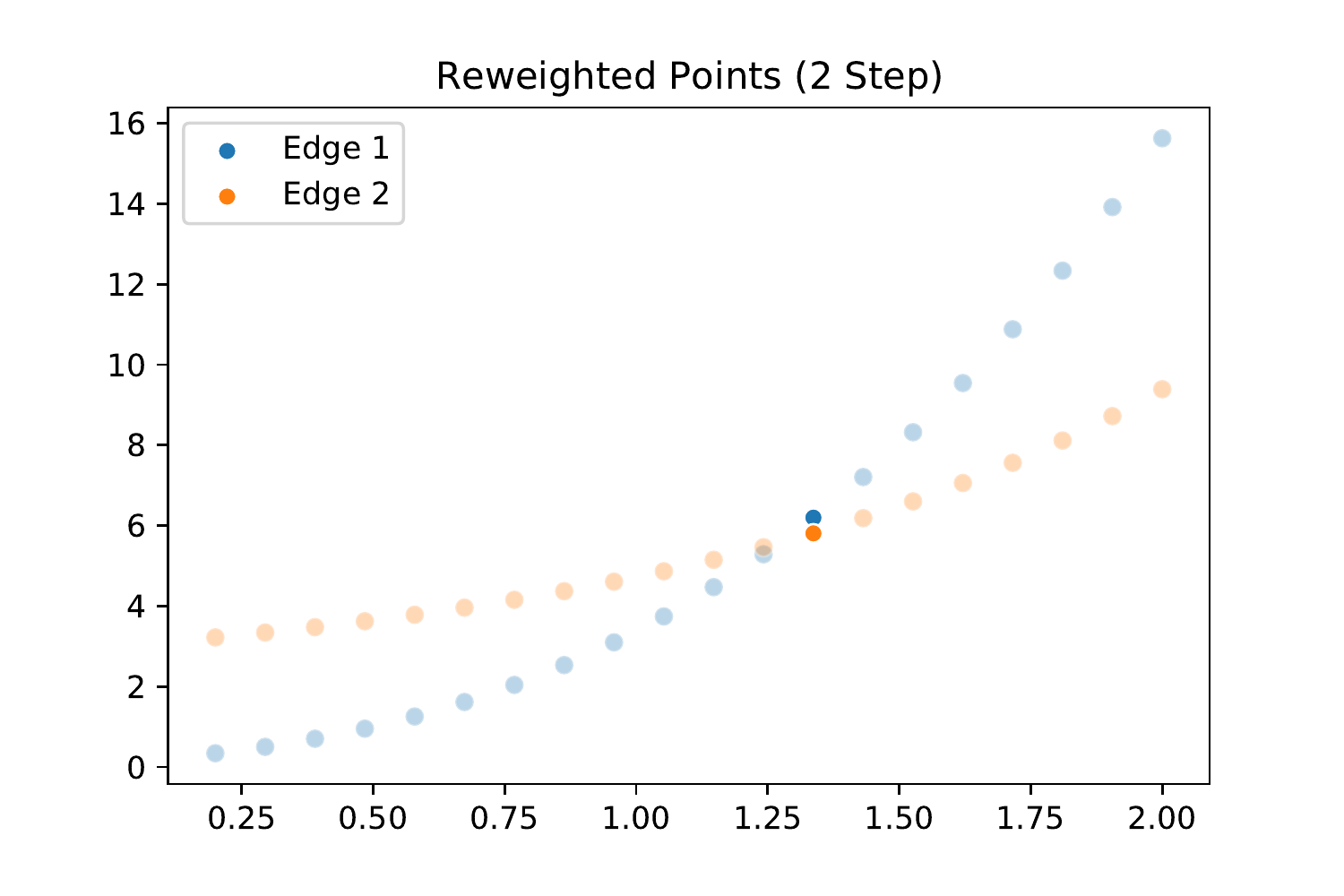}\end{subfigure}
     \begin{subfigure}[t]{0.48\linewidth}\includegraphics[width=\textwidth]{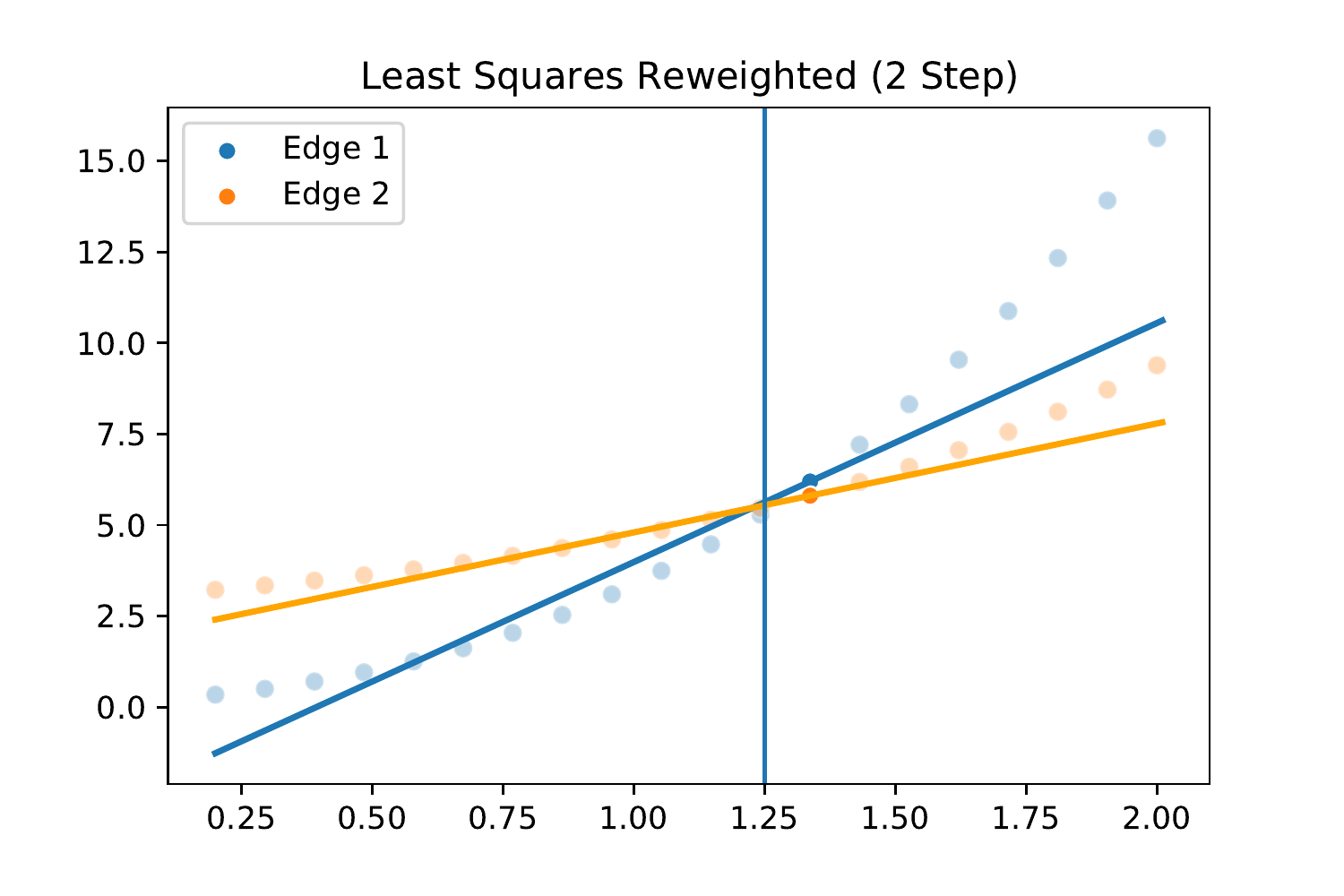}
    \end{subfigure}
    \caption{Step-by-step look at iteratively re-weighting data points for prediction based on task loss. }
    \label{fig:reweightingWalkThrough}
\end{figure}

By re-weighing prediction loss by task loss, our aim is to find a model that can approximate the true decision boundary while remaining more interpretable than the SPO approach. Figure \ref{fig:reweightingWalkThrough} shows a step-by-step look at the reweighing approach. In the left-column we have the weights of each data point, and the right column shows the resulting linear model. We can see that as we iterate between re-weighing data points and generating classifiers using weighted least squares we converge on the true decision boundary. The resulting models also intuitively look more similar to the underlying model than the SPO+ approach.

\section{Experiments} 

An implementation of our experiments can be found here: \url{https://github.com/angelamzhou/opt-milp-taskloss}.

\paragraph{Predict-then-optimize vs. feasible task-loss reweighted MSE. }

\begin{figure}[t]
    \centering
    \begin{subfigure}[t]{0.35\linewidth}\includegraphics[width=\textwidth]{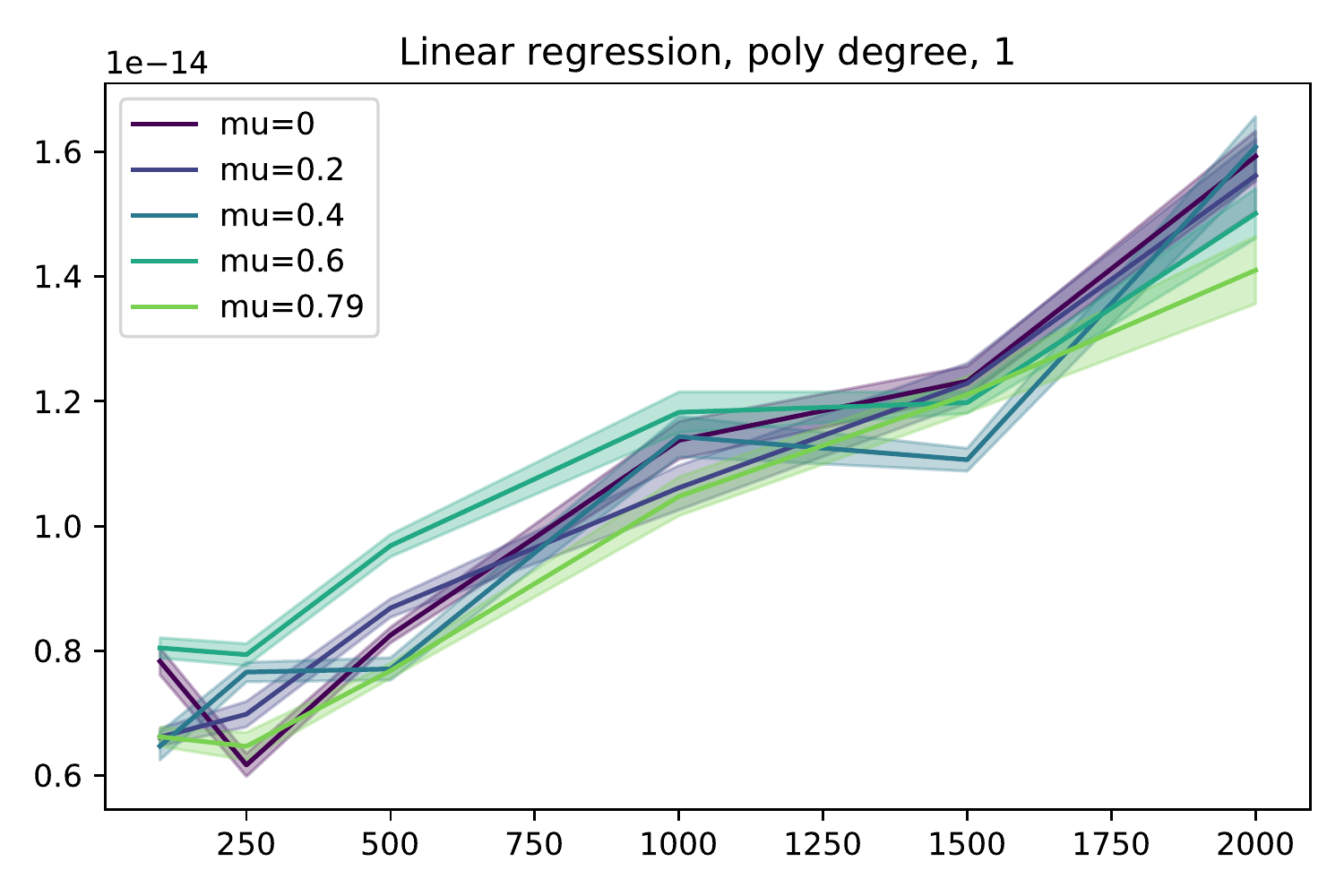}\end{subfigure}\begin{subfigure}[t]{0.35\linewidth}\includegraphics[width=\textwidth]{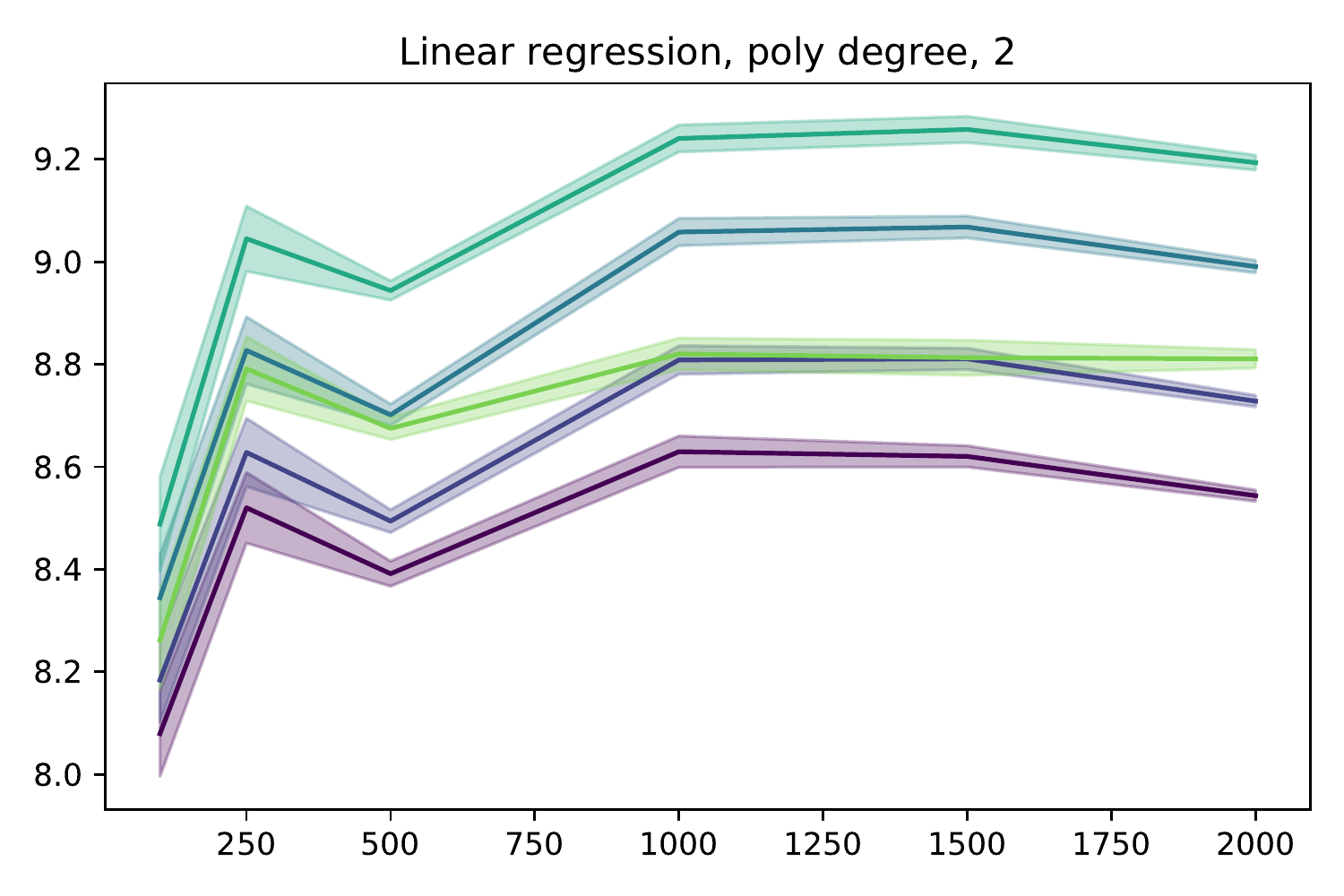}
    \end{subfigure}\begin{subfigure}[t]{0.35\linewidth}\includegraphics[width=\textwidth]{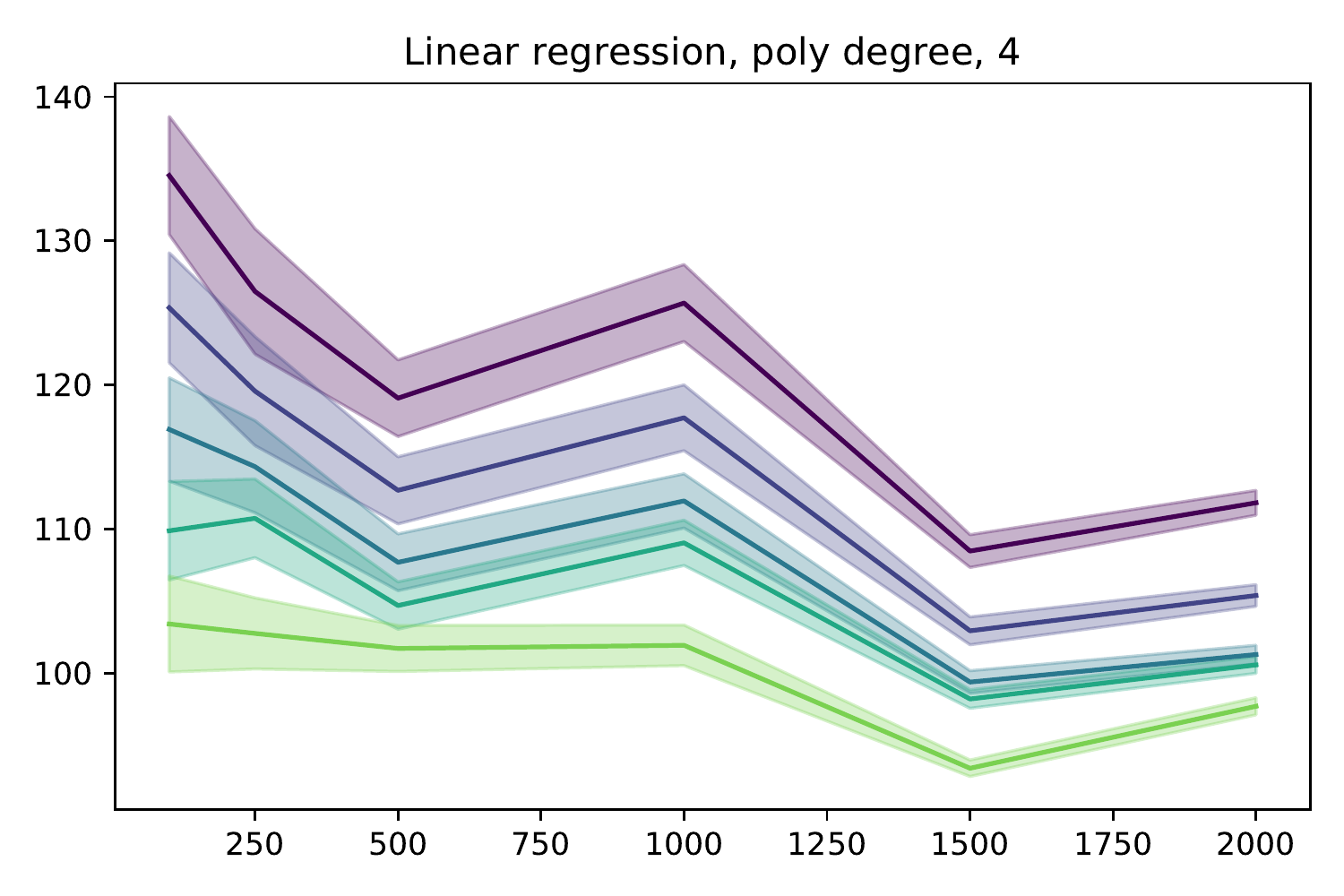}\end{subfigure}
    \begin{subfigure}[t]{0.45\linewidth}\includegraphics[width=\textwidth]{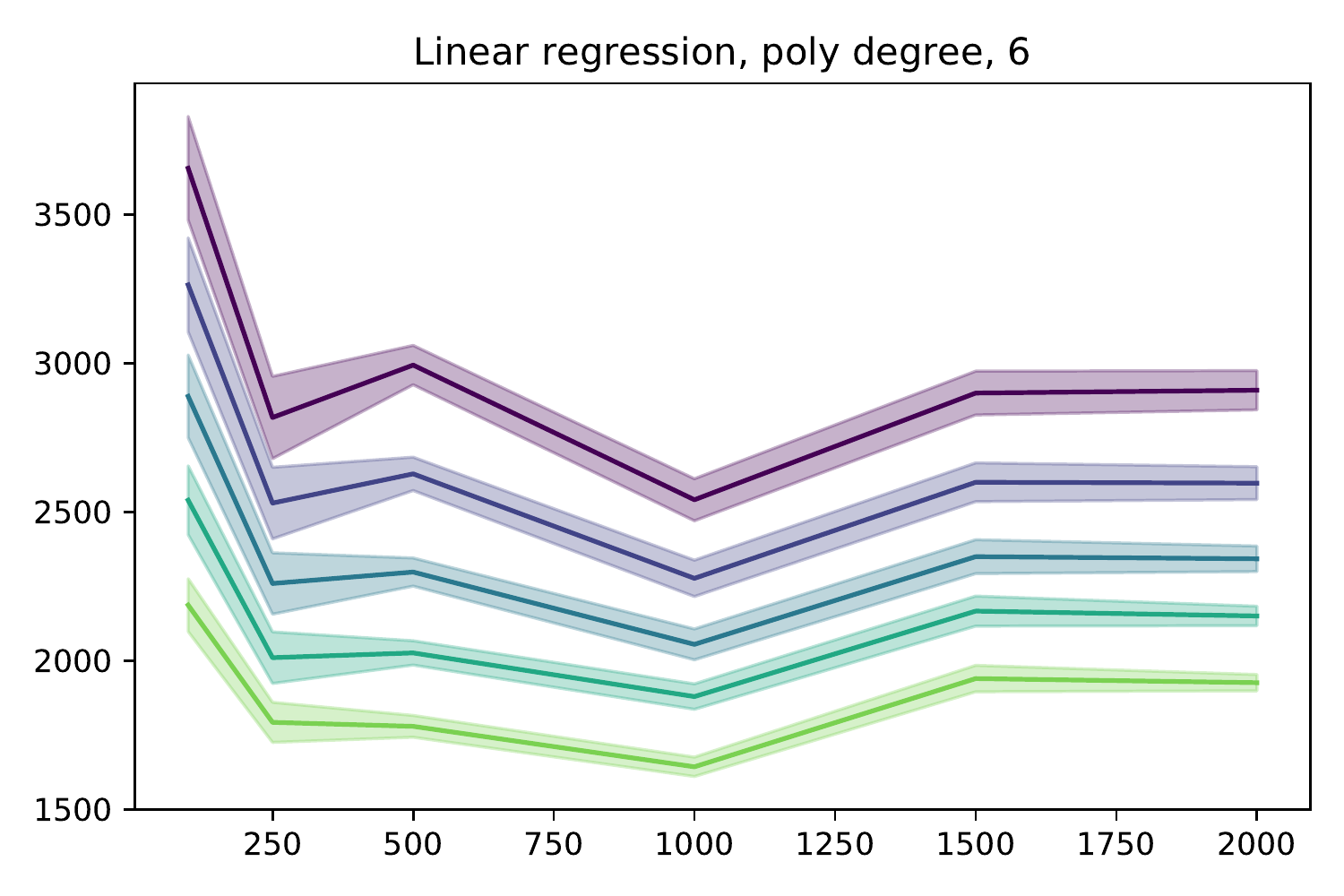}
    \end{subfigure}\begin{subfigure}[t]{0.45\linewidth}\includegraphics[width=\textwidth]{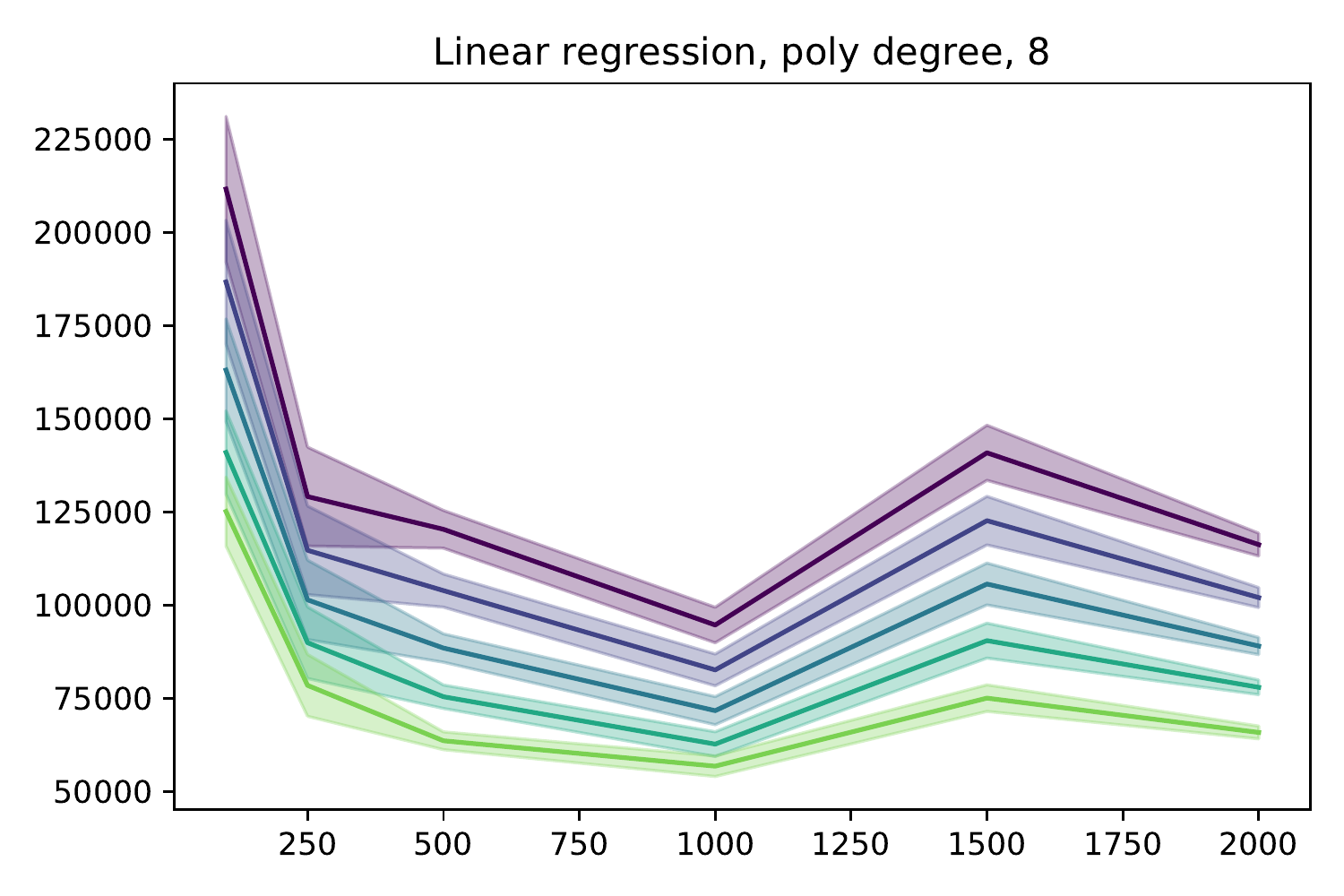}
    \end{subfigure}
    \caption{Following the shortest paths setup of \cite{elmachtoub2017smart}. The regressor is linear regression. Colors indicate training with different mixture weights between $\mu$ times ${ x^*(\hat c) - x^*}$ reweighted MSE vs. $1-\mu$ times prediction MSE. Omitted for same scale are the results for training with $\mu=1$, which is nearly uniformly much worse. }
    \label{fig:reweighted_OLS}
\end{figure}

In Figure \ref{fig:reweighted_OLS} we consider the shortest paths experiment setup of \cite{elmachtoub2017smart}. We set up a network on a 5x5 grid with source node the upper left, and destination the lower right. Edge costs follow a polynomial data-generating process from $\context$ covariates. Specifying the degree of the polynomial increases model misspecification. We investigate Algorithm \ref{alg-decision-reweighted-MSE} with one round of reweighting, with a linear model (ordinary least squares) as the predictor. Because of the zero-inflated nature of the empirical decision regret, we consider an additional hyperparameter $\nu$ which mixes uniform weighting with the empirical decision regret. We plot the decision-regret learning curves as we increase the amount of data used to train the predictors, evaluated on an out-of-sample test set of $N=10000$ independent samples. We plot the achieved test error, averaged over 50 replications (draws of datasets). As we range the hyperparameter from $\nu=0$ (no decision-risk reweighting) to $\nu = 0.8$, the test error decreases for more misspecified models and we achieve sizeable relative regret improvements from 30-60\%, even asymptotically (for large training datasets). (Note the y-axis of the first plot is on the order of $10^{-14}$; linear models achieve near 0 regret if they were in fact well-specified).   

In Figure \ref{fig:reweighted_rf} we repeat the experiment with random forests as a nonparametric predictor. We expect that reweighting does not change the predictions of random forests much, and that random forest prediction is nonparametric so suffers less from misspecification. We find that predict-then-optimize with RF works well, much better than LR as well as better than reweighted LR in general. For highly nonlinear DGP, weighting helps a bit. 

\begin{figure}
    \centering
    \begin{subfigure}[t]{\linewidth}\includegraphics[width=\textwidth]{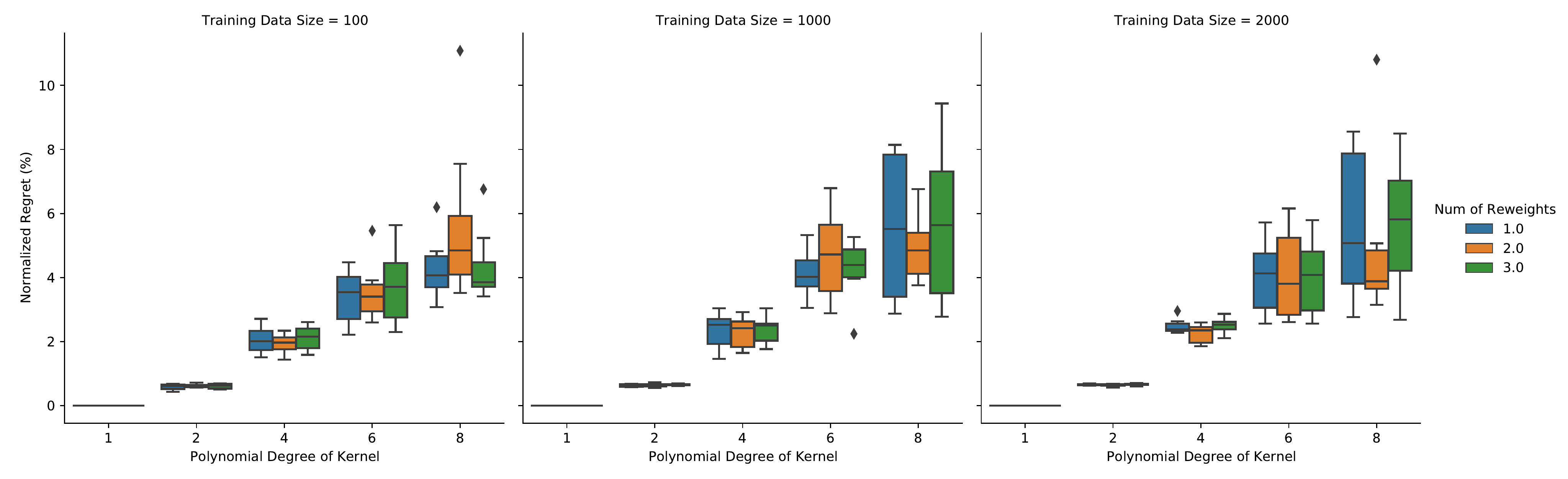}\end{subfigure}
    \caption{Impact of multiple re-weighting iteration on out of sample performance. }
    \label{fig:num_reweighs}
\end{figure}

\paragraph{Effect of repeated re-weightings}
Following the above shortest path experiment, we evaluate the effect of increasing $K$ in algorithm 1. Figure \ref{fig:num_reweighs} show the out of sample normalized regret for a linear regression model after $K \in \{1,2,3\}$ task-based reweighings. The results show that increasing the number of iterations has a negligible effect on the performance of the regressor.

\begin{figure}
    \centering
    \begin{subfigure}[t]{\linewidth}\includegraphics[width=\textwidth]{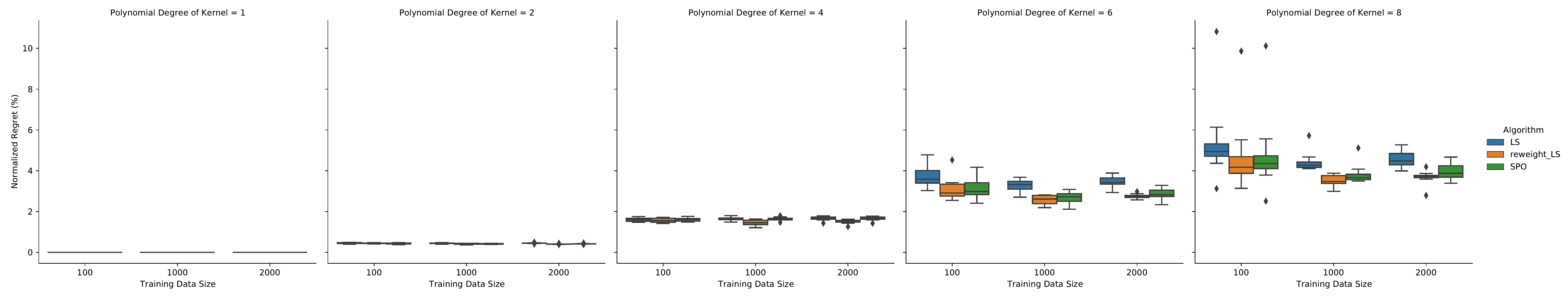}\end{subfigure}
    \caption{Benchmarking against SPO+}
    \label{fig:benchmark}
\end{figure}

\paragraph{Comparison to SPO+} 
We also benchmarked the performance of our approach against SPO+ \cite{elmachtoub2017smart} and evaluated it on the Shortest Path set-up outlined above. Unlike the experiment in the original paper, we implemented an SGD implementation of SPO+, rather than running the reformulated LP. We put a 5 minute time limit on the SGD implementation of SPO+. For our approach (denoted reweighted LS), we ran the experiment with 5 different potential mixture weights and present the mixture weight with the best out of sample performance. For both approaches we use linear regression as the regressor. Figure \ref{fig:benchmark} plots the normalized out of sample approach for each approach. The task loss reweighted predictor performs comparably with SPO+, both of which outperform the predict than optimize approach. 

\section{Conclusion}
We find decision-reweighting the prediction MSE achieves improvements relative to ``predict-then-optimize"; at least for misspecified models. These improvements are competitive with other end-to-end approaches such as SPO+, but very simple to implement. It is a reasonable baseline for contextual linear optimization. 

\paragraph{Acknowledgments}
We thank Priya Donti and Bryan Wilder for helpful discussions. 

\bibliography{prediction-and-opt}
\bibliographystyle{abbrvnat}
\clearpage

\begin{appendix}
\begin{figure}
    \centering
    \begin{subfigure}[t]{0.35\linewidth}\includegraphics[width=\textwidth]{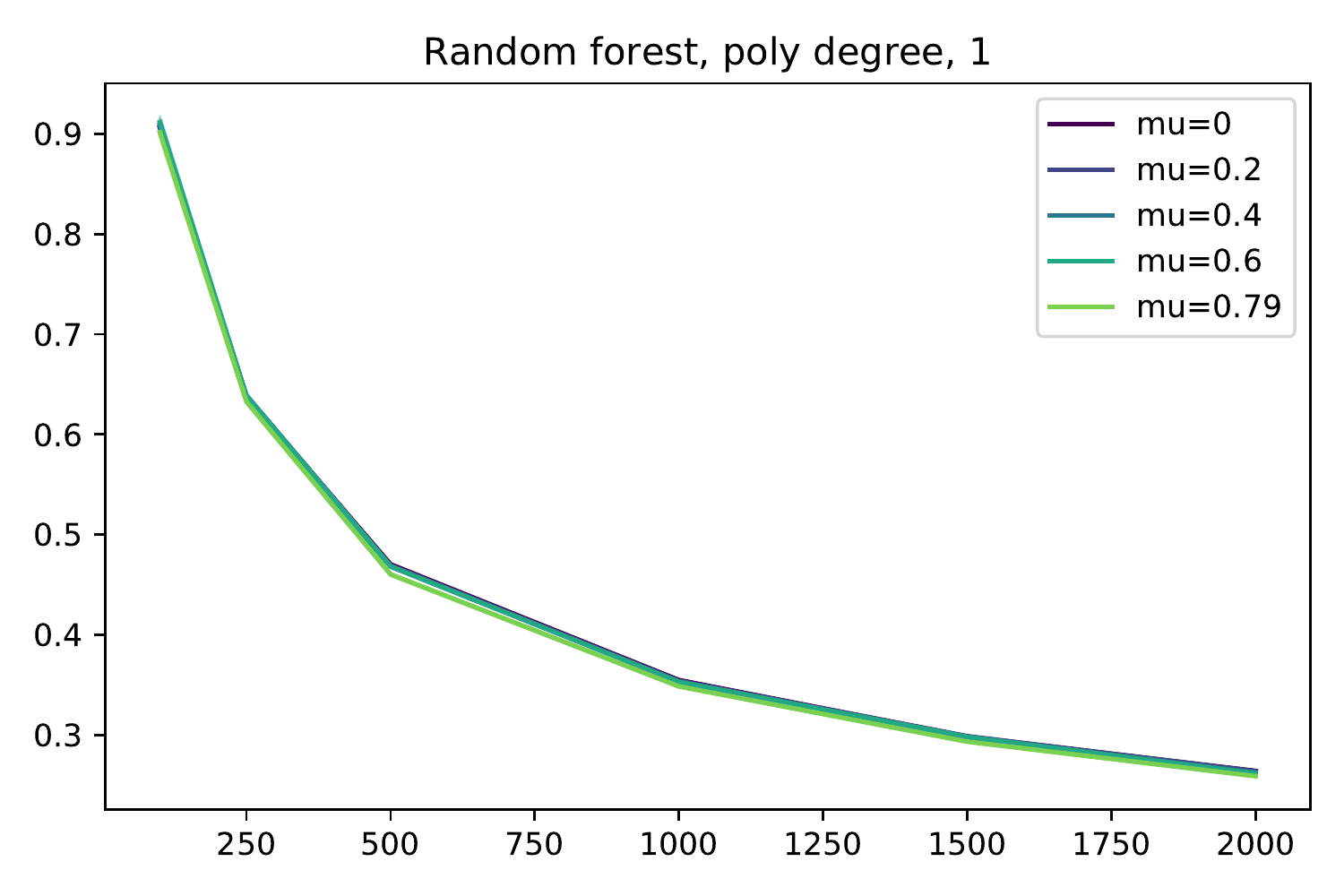}\end{subfigure}\begin{subfigure}[t]{0.35\linewidth}\includegraphics[width=\textwidth]{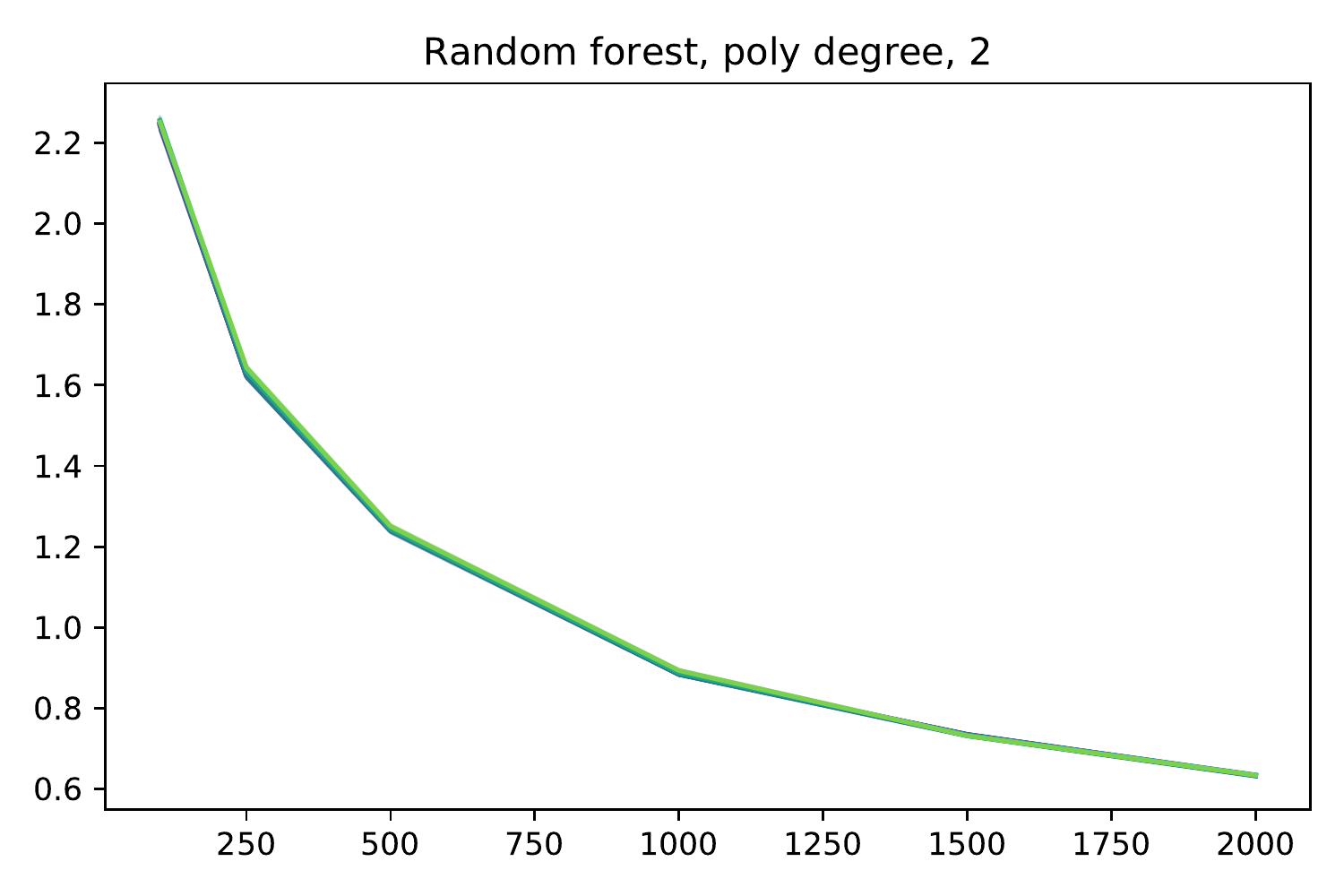}
    \end{subfigure}\begin{subfigure}[t]{0.35\linewidth}\includegraphics[width=\textwidth]{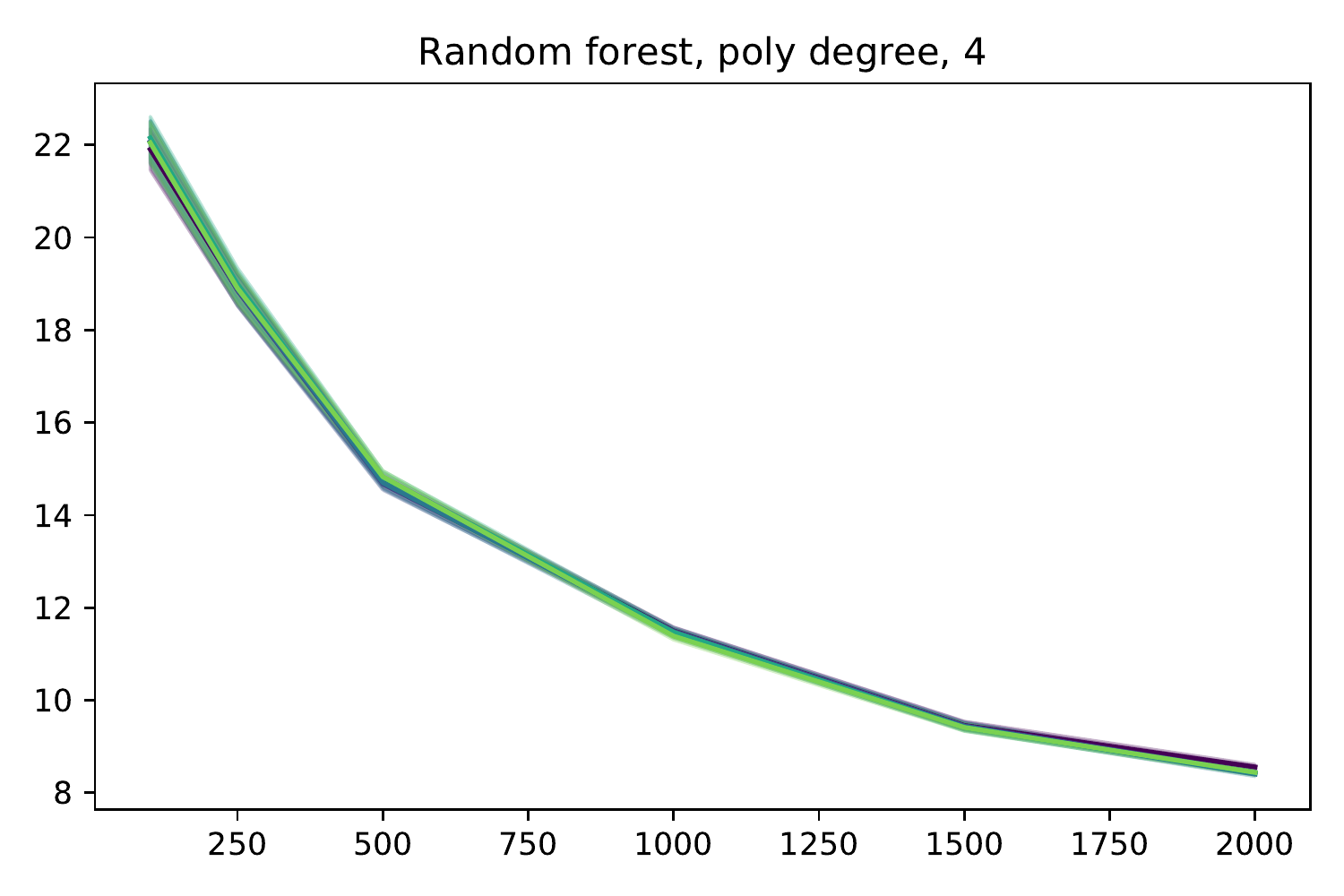}\end{subfigure}
    \begin{subfigure}[t]{0.45\linewidth}\includegraphics[width=\textwidth]{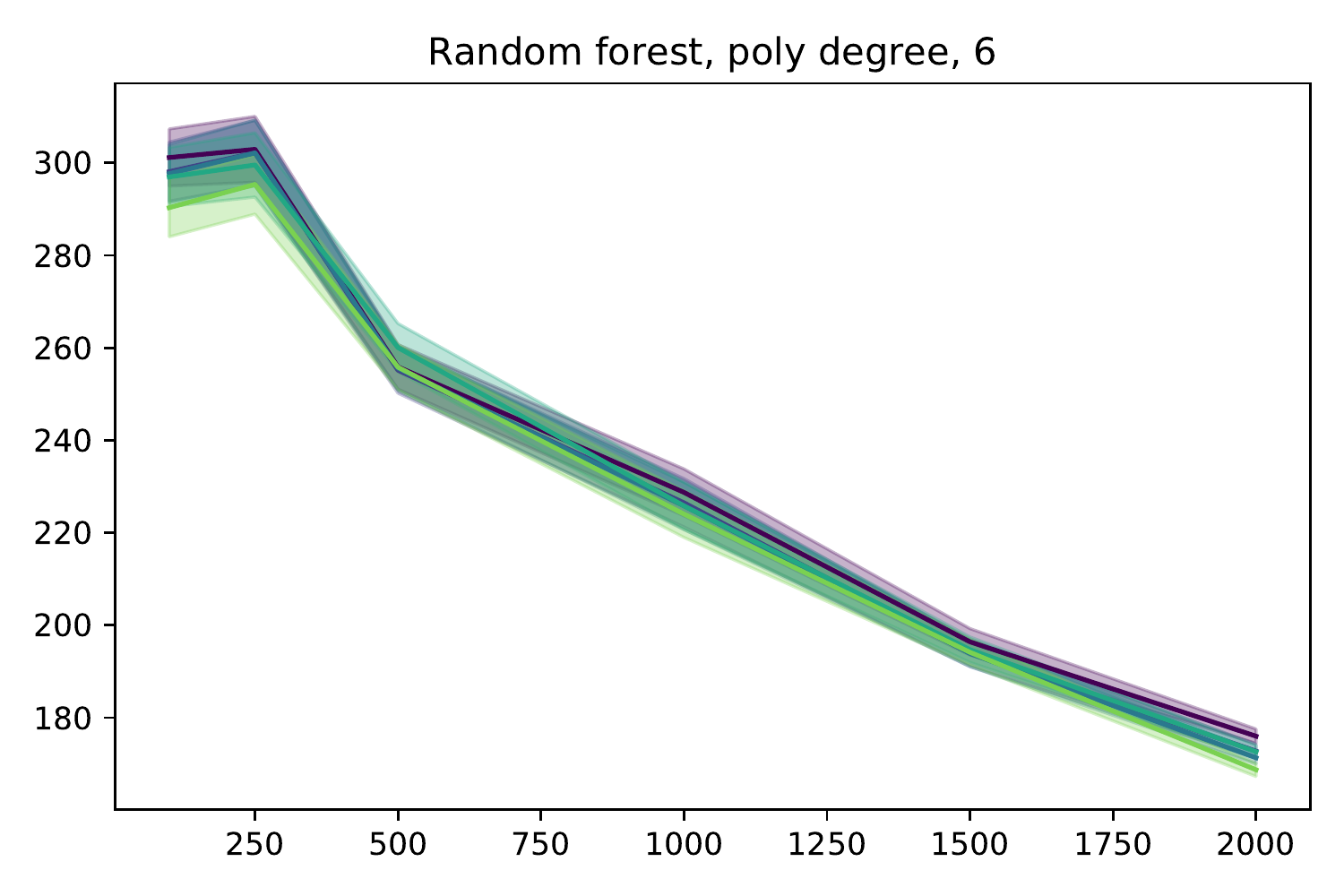}
    \end{subfigure}\begin{subfigure}[t]{0.45\linewidth}\includegraphics[width=\textwidth]{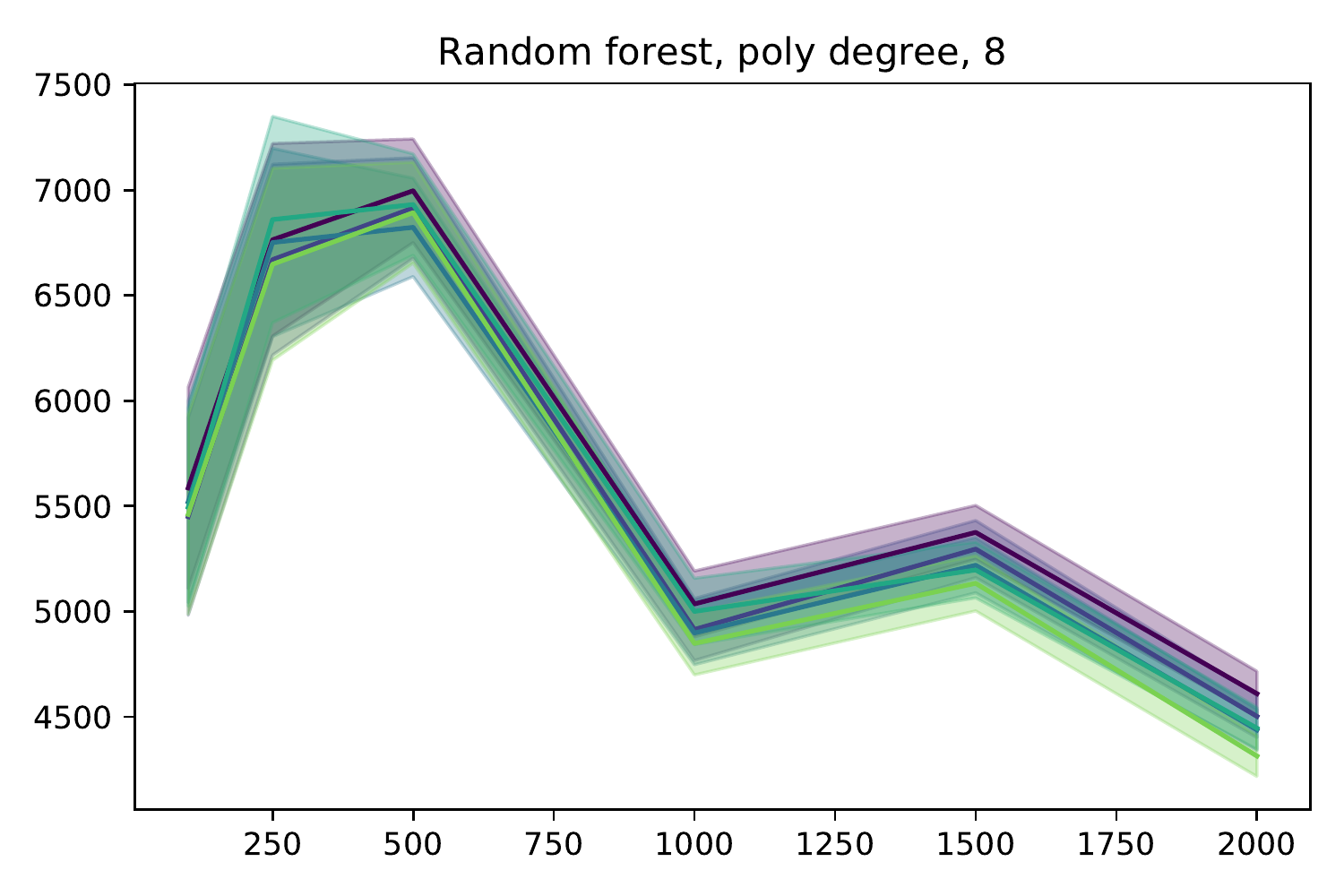}
    \end{subfigure}
    \caption{Following the shortest paths setup of \cite{elmachtoub2017smart}. The regressor is random forest. }
    \label{fig:reweighted_rf}
\end{figure}
\end{appendix}

\end{document}